\documentclass[pdflatex,sn-mathphys-num]{sn-jnl}

\usepackage{graphicx}%
\usepackage{multirow}%
\usepackage{amsmath,amssymb,amsfonts}%
\usepackage{amsthm}%
\usepackage{mathrsfs}%
\usepackage[title]{appendix}%
\usepackage{xcolor}%
\usepackage{textcomp}%
\usepackage{manyfoot}%
\usepackage{booktabs}%
\usepackage{algorithm}%
\usepackage{algorithmicx}%
\usepackage{algpseudocode}%
\usepackage{listings}%

\usepackage{xcolor}%

\theoremstyle{thmstyleone}%
\theoremstyle{thmstyletwo}%

\theoremstyle{thmstylethree}%

\begin{document}

\title[Article Title]{Hybrid Physics-AI Framework of Body Center of Mass Dynamics from Wrist-Worn Sensors}


\author[1]{\fnm{Shuhao} \sur{Que}}\email{s.que@utwente.nl}

\author[2]{\fnm{Valentina} \sur{Breschi}}\email{v.breschi@tue.nl}

\author[1]{\fnm{Ying} \sur{Wang}}\email{imwywk@gmail.com}

\affil[1]{\orgdiv{Electrical Engineering}, \orgname{University of Twente}, \orgaddress{\street{Drienerlolaan 5}, \city{Enschede}, \postcode{7522 NB}, \country{Netherlands}}}
\affil[2]{\orgdiv{Electrical Engineering}, \orgname{Eindhoven University of Technology}, \orgaddress{\street{Groene Loper 19}, \city{Eindhoven}, \postcode{5612 AP}, \country{Netherlands}}}


\abstract{Wrist-worn wearables, such as smartwatches, have become people's accessory to wear on a daily basis, which usually incorporates an inertial measurement unit (IMU). Wrist-worn IMU has been widely used to assess physical activity and estimate energy expenditure, forming the basis of daily-life health monitoring technologies. Yet, it does not fully represent whole-body dynamics, for which the body center of mass (COM) is considered the physiological reference standard. Therefore, this work proposes a simplified kinematic model (KM) based on kinematic equations used for robotic arm control. The simplified KM model is designed in this work to map the wrist IMU to the COM acceleration. It is built upon several reductive assumptions that enable the solvability of the dynamic equations based on wrist IMU measurements alone. This work further proposes three types of hybrid AI modeling methods, namely human kinematic model-based neural network (HKM-NN) models, to leverage the power of both grey-box modeling and black-box modeling. The NN component is based on two types of NN architectures, a fully-connected neural network (FCNN) and a long-short-term memory layer (LSTM). The HKM-NN methods include serial learning and two approaches of simultaneous learning. The former trains the KM and NN components sequentially, while the latter two train them simultaneously, but with distinct coupling strategies in the loss function. The proposed models are trained and tested using our dataset, which includes wrist IMU measurements and ground-truth COM measurements from 10 healthy volunteers during six gait activities (i.e., natural walking, narrow beam walking, uneven terrain walking, slalom walking, upstairs and downstairs walking, and walk and turn) and sit-to-stand (SS) transitional movement. The results demonstrate the feasibility of estimating COM acceleration from wrist IMU measurements. Our KM model yields satisfactory results, with a normalized error ranging from 6.7\% to 12.5\% for gait activities and a normalized error of 5.6\% for the SS activity. In comparison with the KM model, our hybrid AI: HKM-NN models significantly enhance the performance, achieving 5.3\% to 9.3\% normalized errors for gait activities, and the best normalized error of 3.9\% for the SS activity. In addition, the HKM-NN models demonstrate distinct robustness characteristics under noisy test conditions, with the simultaneous learning framework generally maintaining greater robustness under Gaussian perturbations, while the KM model exhibits comparatively strong robustness under salt-and-pepper noise. These findings highlight the importance of combining biomechanical structure with data-driven learning for wearable sensing applications operating under imperfect and noisy measurement conditions.}




\maketitle

\section{Introduction}\label{sec1}
Nowadays, consumer wrist-worn wearables (e.g., smartwatches, fitness trackers, etc.) have become a common accessory for daily use. These devices usually incorporate an inertial measurement unit (IMU) that records both linear accelerations (accelerometer) and angular velocities (gyroscope). Because of their comfort and social acceptability, wrist-worn IMUs are highly practical for daily-life monitoring and have therefore been widely investigated for physical activity recognition~\cite{larsen2022effectiveness} and energy expenditure estimation~\cite{pope2019accuracy, fuller2020reliability}. However, IMU signals from the wrist reflect mainly local arm movement and do not represent the motion of the whole body, which limits their predictive power in these applications~\cite{larsen2022effectiveness, fuller2020reliability}.

A key to unlocking their full potential lies in estimating the whole-body dynamic descriptor, i.e., the acceleration of the body’s center of mass (COM), from the local measurements. COM acceleration serves as a key biomechanical descriptor of dynamic stability and functional mobility during body motion~\cite{winter2009biomechanics}. Reliable estimation of COM acceleration in daily life offers a compact, physically-grounded representation of whole-body dynamics, capturing how people maintain balance, move efficiently, and adapt to everyday challenges. However, the gold-standard method to estimate COM acceleration relies on an opto-electronic system (e.g., VICON) with markers attached to the human body~\cite{b6} or a full-wear body suit embedded with IMUs, e.g., Xens MVN Link motion capture suit. Since the former confines subjects to a limited laboratory space and the latter provides portability at the cost of wearability, both approaches are unsuitable for long-term daily-life monitoring. While IMUs positioned near the pelvis or lumbar region can approximate COM dynamics~\cite{b11, b12, b14, kerns2023effect, fino2020inertial, b5}, their placement remains intrusive for daily-life use. Similarly, pocket-based inertial sensing suffers from limited practicality, as smartphones are frequently removed from the pocket during daily activities, while carrying a dedicated sensor solely for monitoring introduces additional burden and inconsistent placement. In contrast, the wrist is one of the most socially accepted and universally worn sites for inertial sensing due to the widespread adoption of smartwatches, making it an ideal candidate for scalable digital health monitoring in daily life.

However, translating wrist movement into COM dynamics remains an unsolved computational challenge because this is a fundamentally ill-posed inverse problem, i.e., multiple whole-body motions can give rise to similar wrist IMU signals, making the mapping from wrist movement back to COM dynamics non-unique and highly sensitive to measurement noise and distributional variability encountered in real-world wearable sensing. Two methodological paradigms can be applied to such inverse problems: 1) a physics-based modeling approach, which formulates explicit biomechanical equations to estimate COM motion from segmental kinematics~\cite{sylvester2021review, mathieu2023biomechanical}. While such models yield interpretable and physiologically grounded results, they typically rely on full-body measurements, often requiring up to 17 IMUs distributed across the body, to ensure solvability, which restricts their use in daily-life scenarios. In addition, physics-based models' performance depends critically on the validity of model assumptions and input quality. In practice, such models can be sensitive to measurement noise when operating under reduced sensing configurations (i.e., only 1 IMU available on the wrist); 2) a data-driven learning approach, where neural networks are trained to approximate the nonlinear mapping between wrist-worn IMU signals and COM acceleration. Although these models could, in principle, operate with limited sensing, no studies have yet demonstrated their effectiveness for COM estimation from a single wrist-worn IMU. Moreover, their robustness depends on the underlying architecture and training distribution. For example, models without temporal structure may be highly sensitive to noise~\cite{zhang2016understanding}, while even more advanced models can degrade under data distribution shift due to a lack of physical constraints~\cite{karniadakis2021physics}. Therefore, addressing this ill-posedness requires mechanisms to constrain the solution space and stabilize the estimation, while accounting for both the structural limitations of physics-based models and the data dependence of learning-based approaches. This motivates the integration of biomechanical priors or physical constraints within data-driven frameworks. Such integration can take different forms, such as sequential strategies that incorporate physics-based estimates into subsequent data-driven learning or more tightly coupled approaches in which physics-based and neural components interact within the learning process. However, the impact of different coupling strategies on robustness and generalization remains unclear. This highlights a critical but underexplored challenge: how different strategies of integrating physical knowledge and learning-based models in a principled manner influence stability, robustness, and generalization in minimal-sensing COM estimation.

Accordingly, we propose a human kinematic model–based neural network (HKM-NN) framework that integrates a simplified human kinematic model with neural networks (FCNN or LSTM) to combine physical interpretability and data-driven adaptability. We propose a simplified human kinematic model (KM), which encodes the coupling between wrist motion and COM acceleration through a reduced set of biomechanical equations. The proposed KM model is rendered computationally solvable within a reduced-order formulation and tailored to locomotor and transitional activities under reductive assumptions. Three physics–AI integration schemes, i.e., serial learning and two variants of simultaneous learning, are designed to examine how assigning the leading role to either the neural network or the physical model influences the interplay between data-driven adaptation and prior biomechanical knowledge in an ill-posed setting. In the physics-leading (serial learning) scheme, the biomechanical model provides the primary physical representation, while data-driven learning introduces additional flexibility to account for dynamics beyond the model assumptions. In the data-driven-leading (simultaneous learning) scheme, the neural network and KM are optimized jointly through a coupled loss. Two variants are investigated. The first approach imposes stronger biomechanical constraints on the learning process, thereby restricting the flexibility of data-driven parameter identification. In contrast, the second adopts a more flexible formulation in which the physics model serves as structural guidance rather than a strict constraint, allowing greater adaptation to the observed data.

\section{Results}\label{sec2}
A total of 10 human subjects participated in this study, who performed locomotor and transitional activities, including six gait activities and sitting-standing (SS) transition activity. The six gait activities include natural walking (NW), narrow beam walking (BW), slalom walking (SW), walk-turn-walk (WT), uneven terrain walking (TW), and upstairs and downstairs walking (UD). The dataset details and protocol are described in ``Dataset" in Methods (see Figure~\ref{fig:gaits}).

Since the medio-lateral (z-) acceleration consistently exhibits the largest dynamic range and root-mean-square values across all activities, it represents the dominant component of COM acceleration. For the detailed derivation on axis dominance order, see ``Axis dominance order" in the Supplementary. Consequently, performance on the z-axis serves as the most informative axis-specific evaluation and is used for the following result analyses. In addition to the dominant z-axis, the acceleration magnitude is also evaluated to assess how well the models preserve the resultant three-dimensional acceleration vector, which is particularly relevant for downstream biomechanical analyses.

\subsection{Baseline and HKM-NN models comparison}
For comparison with the HKM-NN models, two categories of baseline models are implemented: the KM model and NN-based models, including FCNN and LSTM. The FCNN consists of two fully connected layers with a hidden size of 16, while the LSTM model employs a hidden size of 8. Three hybrid learning strategies are investigated to combine the KM and NN models: serial learning (ser-), simultaneous learning with ground-truth supervision on the KM (sim1-), and simultaneous learning with an unsupervised KM prior (sim2-). The ser- approach trains the KM and NN models sequentially. In contrast, both simultaneous approaches train the two models jointly using a coupled loss term. In sim1-, the KM is additionally supervised by the ground-truth COM acceleration, whereas in sim2-, the KM is not directly supervised by the ground truth. Instead, it serves as a structural prior that guides the NN through the coupled loss. Specifically, the coupled loss imposes a soft constraint that encourages agreement between the predictions of the KM and NN models. In total, six HKM-NN models are implemented: sim1-LSTM, sim2-LSTM, sim1-FCNN, sim2-FCNN, ser-LSTM, and ser-FCNN (see ``Human kinematic model-based neural networks” in Methods for details).

Three evaluation metrics are used, including max-min normalized root-mean-squared error (NRMSE), coefficient of determination ($R^2$), and Pearson's correlation coefficient ($r$). The NRMSE is calculated by dividing the RMSE by the maximum and minimum value range of the reference COM acceleration. We multiplied the NRMSE value by 100 to derive the error percentage. Detailed calculations are presented in ``Evaluation metrics" in Methods.

Comparisons among the baseline models and HKM-NN models under different signal-to-noise ratio (SNR) conditions of the test data, in terms of $R^2$, are shown in Fig.~\ref{fig:r2_gaussian} (Gaussian noise) and Fig.~\ref{fig:r2_pepper} (salt-and-pepper noise). Additional summarised results based on NRMSE and $r$ are provided in Supplementary Fig.~2, 3, 4, and 5. Detailed reasoning and configuration of the added noise are presented in ``Influence of artificial noise" in Methods.

\subsubsection{Noise-free test data}
The summarized results based on NRMSE and $R^2$ are shown in Figure~\ref{fig:models_nrmse} and Figure~\ref{fig:models_r2}, respectively. The results based on $r$ are provided in Supplementary Fig.~1 as $r$ shows limited differences in performance across the models.

Overall, the NN-based and most HKM-NN models demonstrate broadly comparable performance across activities in terms of both NRMSE and $R^2$, with no statistically significant difference. The only exception is the sim2-based models, which underperform not only consistently compared to other NN and hybrid models, but also in many cases, the standalone KM model. Such performance inferiority is especially observed for the acceleration magnitude. One possible explanation is that, in the absence of ground-truth supervision, the KM lacks a reliable optimization objective and may converge to suboptimal solutions, which subsequently limits the effectiveness of the structural guidance provided to the NN. Apart from the sim2 variants, the KM model generally exhibits slightly larger NRMSE values and lower $R^2$ values than the NN-based and HKM-NN models, indicating comparatively reduced predictive accuracy and limited flexibility in capturing activity-dependent acceleration dynamics.

\begin{figure}[H]
    \centering
        \centering
        \includegraphics[width=\textwidth]{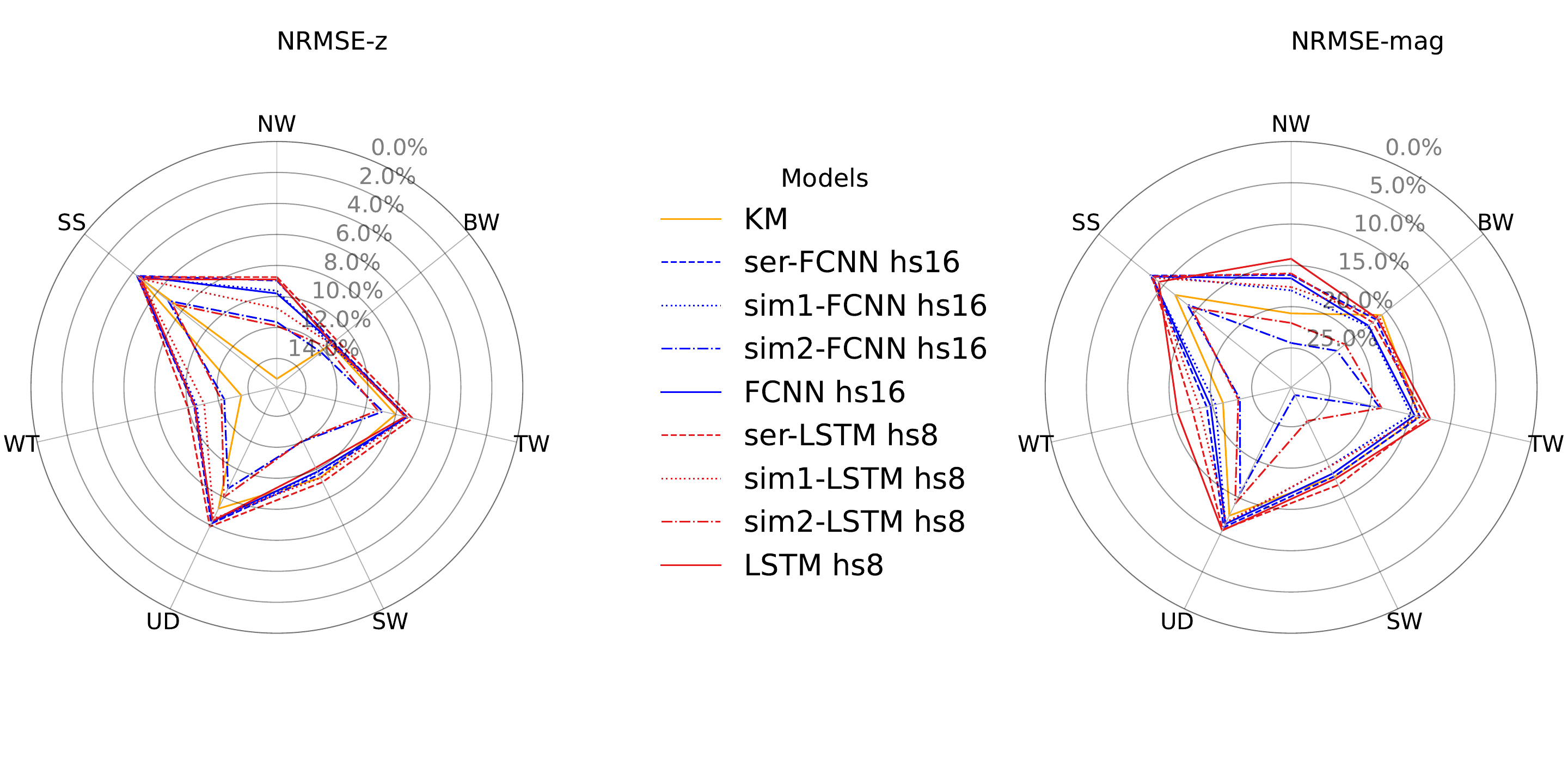}
        \begin{minipage}{0.8\linewidth}
            \footnotesize
            KM: kinematic model; sim1-/sim-2: simultaneous learning; ser-: serial learning; \\
            NW: natural walking; BW: beam walking; TW: terrain walking; SW: slalom walking; UD: upstairs and downstairs walking; WT: walk and turn; SS: sitting and standing up.
        \end{minipage}
        
        \caption{Max–min normalized root-mean-square error (NRMSE) computed for the z-axis and the acceleration magnitude (mag), representing the error as a percentage of the ground-truth value range. The z-axis denotes the medio-lateral direction.}
        \label{fig:models_nrmse}
\end{figure}

\begin{figure}[H]
        \centering
        \includegraphics[width=\textwidth]{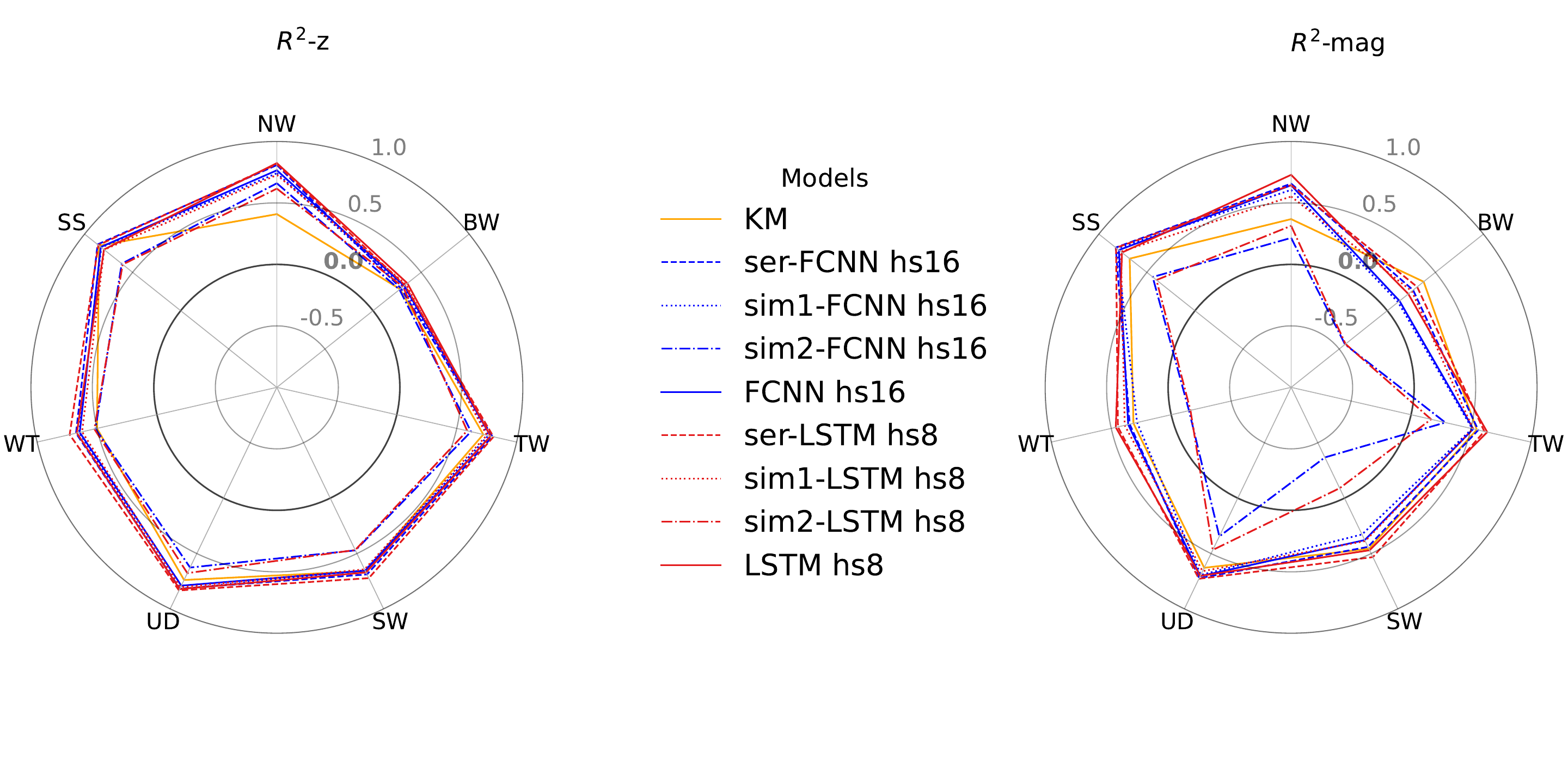}
        \begin{minipage}{0.8\linewidth}
            \footnotesize
            KM: kinematic model; sim1-/sim-2: simultaneous learning; ser-: serial learning; \\
            NW: natural walking; BW: beam walking; TW: terrain walking; SW: slalom walking; UD: upstairs and downstairs walking; WT: walk and turn; SS: sitting and standing up.
        \end{minipage}
        
        \caption{Coefficient of determination ($R^2$) along the z-axis and the acceleration magnitude (mag). The z-axis denotes the medio-lateral direction.}
        \label{fig:models_r2}
\end{figure}

\subsubsection{Gaussian-noise test data}
\begin{figure}[H]
    \centering
    \includegraphics[width=\linewidth]{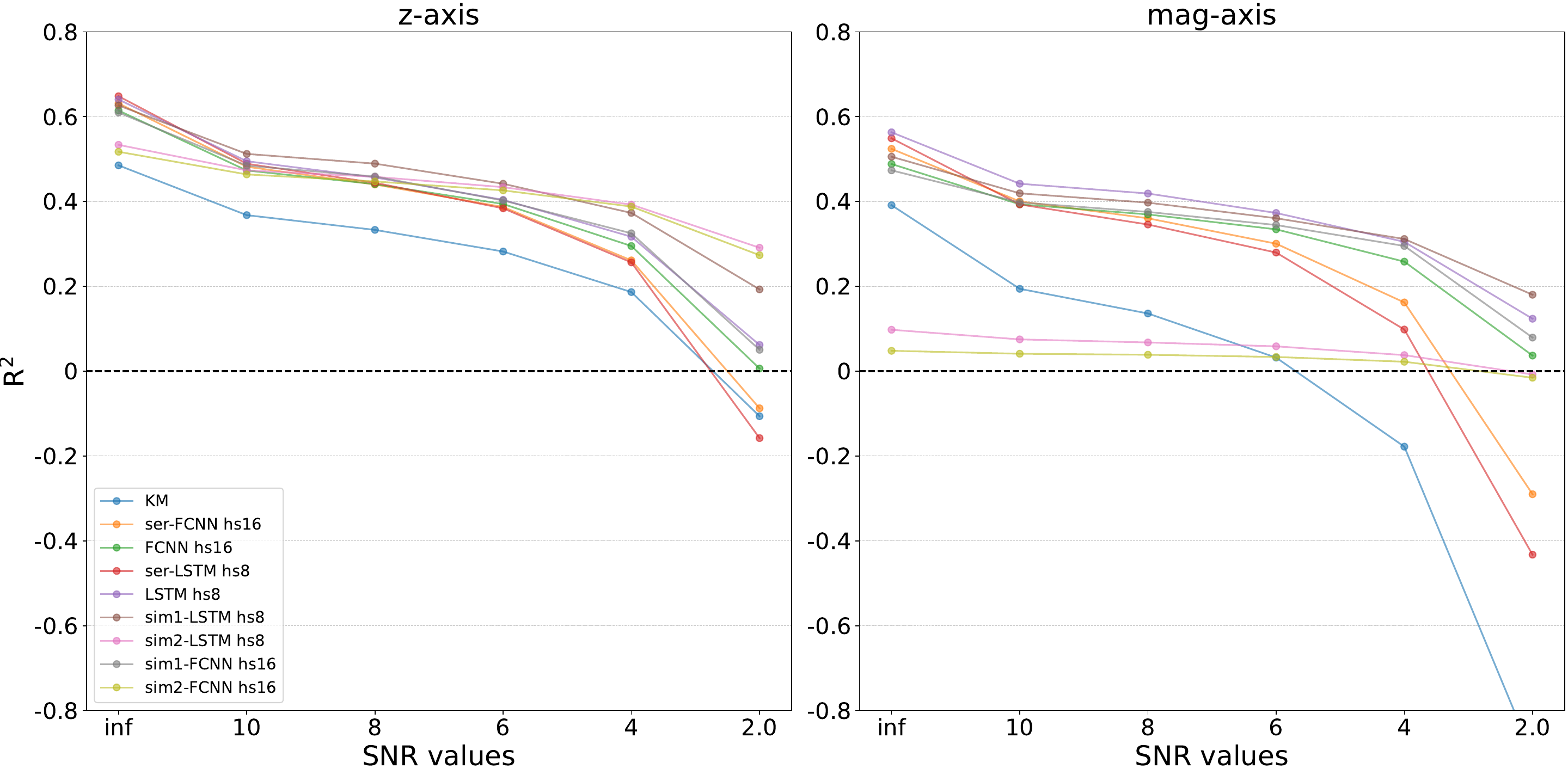}
    \caption{Coefficient of determination ($R^2$) along the z-axis and the acceleration magnitude under different Gaussian signal-to-noise ratio (SNR) conditions of the test data, including inf (i.e., no noise), 10, 8, 6, 4, and 2. The z-axis corresponds to the medio-lateral direction.}
    \label{fig:r2_gaussian}
\end{figure}

\begin{figure}[H]
    \centering
    \includegraphics[width=\linewidth]{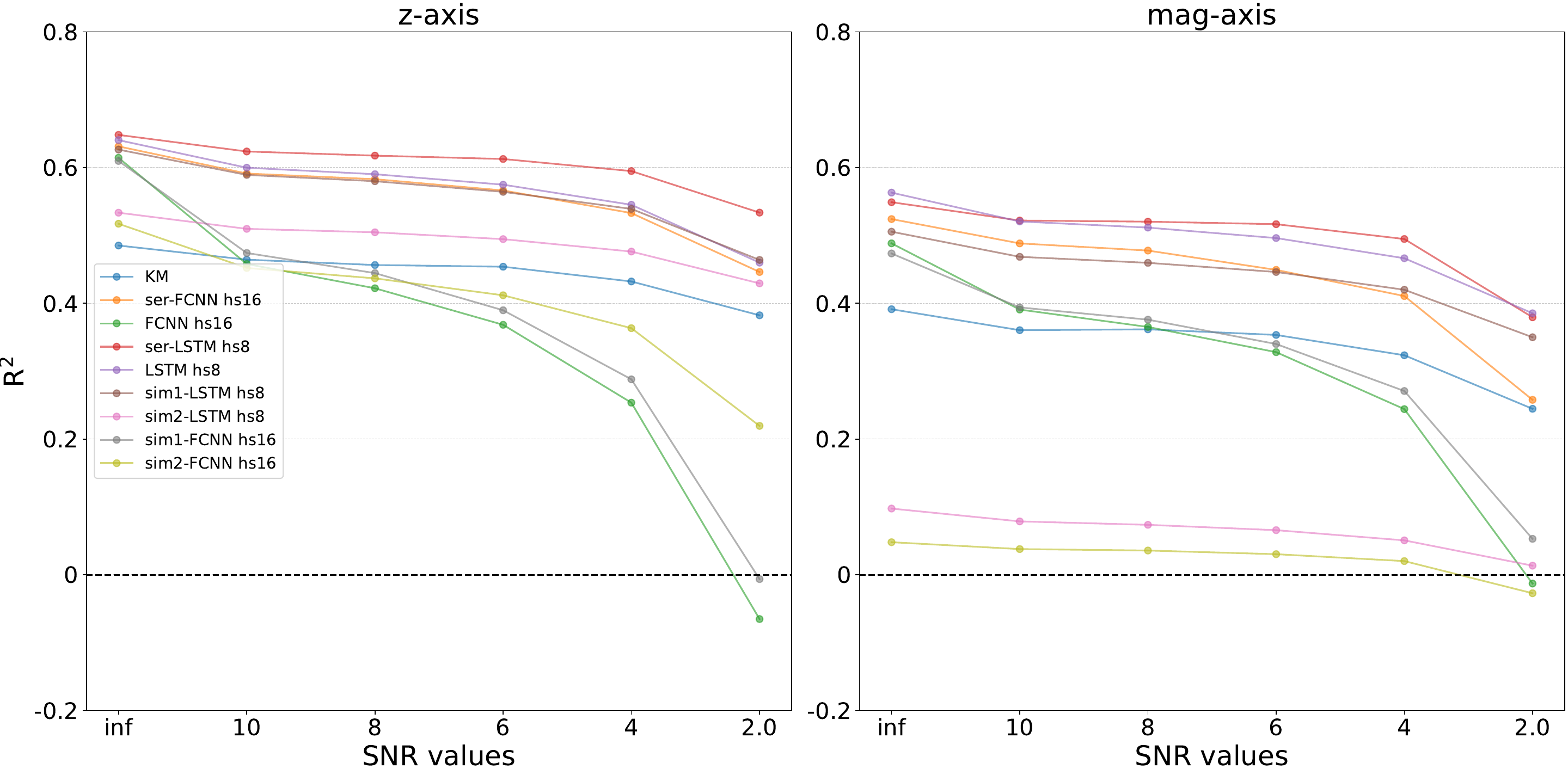}
    \caption{Coefficient of determination ($R^2$) along the z-axis and the acceleration magnitude under different salt-and-pepper signal-to-noise ratio (SNR) conditions of the test data, including inf (i.e., no noise), 10, 8, 6, 4, and 2. The z-axis corresponds to the medio-lateral direction.}
    \label{fig:r2_pepper}
\end{figure}

The models' performance based on $R^2$ under varying Gaussian SNR conditions of the test data is shown in Figure~\ref{fig:r2_gaussian}. Overall, the predictive performance of all models gradually deteriorate as the SNR decreases, indicating the increasing difficulty of acceleration estimation under stronger noise perturbations. However, substantial differences in robustness are observed among the models. 

The standalone KM model exhibits the greatest sensitivity to Gaussian noise, with its $R^2$ values decreasing rapidly as the SNR decreases and becoming strongly negative under the most severe noise conditions. This suggests that the deterministic biomechanical formulation of the KM model lacks sufficient flexibility to compensate for dense stochastic perturbations introduced by Gaussian noise. The ser-based models (ser-FCNN and ser-LSTM) also exhibit substantial performance degradation under decreasing SNR conditions, particularly at low SNR values. Since the ser- framework relies on the output of the KM model as an intermediate representation, the Gaussian noise sensitivity of the KM model is likely propagated into the subsequent NN model. Consequently, errors introduced by the KM component may accumulate and negatively affect the downstream learning process. 

The two simultaneous learning strategies exhibit distinct responses to Gaussian noise. For the z-axis, the sim2-based models show relatively limited performance degradation as the SNR decreases, despite their comparatively lower performance under the noise-free condition. A similar pattern is observed for the acceleration magnitude, where the sim2-based models maintain consistently lower $R^2$ values but exhibit relatively small changes across SNR conditions. This suggests that the sim2-based models are comparatively insensitive to increasing Gaussian perturbations, although such insensitivity does not translate into greater predictive accuracy. In contrast, the sim1-based models generally maintain higher performance across SNR conditions, particularly for the acceleration magnitude, while also showing relatively gradual performance degradation. Therefore, the two simultaneous strategies appear to exhibit different performance–robustness characteristics. Specifically, sim1- combines relatively high predictive accuracy with robustness to Gaussian perturbations, whereas sim2- exhibits lower predictive accuracy but a comparatively constrained response to increasing noise. The mechanism underlying the reduced noise sensitivity of sim2- cannot be determined from the present results and requires further investigation.

Similar results are observed for NRMSE (see Supplementary Fig.~2). As for $r$ (see Supplementary Fig.~3), all models exhibit comparable performance degradations, with the standalone KM model yielding the lowest $r$ values across all SNR values.

\subsubsection{Salt-and-pepper-noise test data}
The models' performance based on $R^2$ under varying salt-and-pepper SNR conditions of the test data is shown in Figure~\ref{fig:r2_gaussian}. Overall, the predictive performance of most models is less affected by salt-and-pepper noise than by Gaussian noise, with the exceptions including FCNN, sim1-FCNN, and sim2-FCNN models.

Specifically, compared with the Gaussian noise experiments, the KM model demonstrates substantially improved robustness under salt-and-pepper noise conditions, maintaining relatively stable $R^2$ values across decreasing SNR levels. This suggests that the deterministic biomechanical structure of the KM model is less sensitive to sparse impulsive perturbations, as the overall kinematic motion trend remains preserved despite occasional corrupted samples.

The FCNN, sim1-FCNN, and sim2-FCNN models exhibit larger performance degradation under salt-and-pepper noise than their LSTM-based counterparts. This difference may be associated with the architectural limitations of the FCNN in modeling temporal dependencies. As salt-and-pepper noise introduces sparse and abrupt perturbations at individual time points, the FCNN directly maps the locally corrupted input features to the output without explicitly exploiting temporal context. In contrast, the recurrent structure of the LSTM incorporates information across consecutive time steps, which may reduce its sensitivity to isolated impulsive disturbances. The greater degradation observed consistently across the FCNN-based models therefore suggests that temporal modeling plays an important role in robustness to sparse input perturbations. In contrast, the ser-FCNN model demonstrates considerably greater robustness, indicating that the serial integration with the KM may partially compensate for this limitation of the FCNN architecture.

Similar results are observed for both NRMSE (see Supplementary Fig.~4) and $r$ (see Supplementary Fig.~5).

\subsection{4-segment KM model analysis}
The 4-segment KM model comprises 27 parameters, including 3 rotation vectors ($R^1_0, R^2_1, R^3_2 \in \mathbb{R}^4$), 3 relative position vectors ($P^1_0, P^2_1, P^3_2 \in \mathbb{R}^3$), and 2 angular velocity coupling vectors for the angular velocity propagation between segments ($A^1_0, A^2_1 \in \mathbb{R}^3$). The angular velocity coupling vectors are one of the assumptions made by our simplified KM model, detailed in `4-segment (KM) model' in Methods. The KM model's parameter identification and influence evaluation methods are detailed in ``KM model identification" and ``KM model parameter analysis" in Methods.

\subsubsection{Parameter influence}
To evaluate the rank of the Jacobian matrix with respect to model parameters, we calculate the rank at each training epoch. The rank of the model parameter Jacobian matrix is mostly between 16 and 20 (see Supplementary Fig.~7), smaller than 27 (i.e., a full-rank). This indicates that some parameters of our 4-segment KM model are redundant, or that there is collinearity between parameters, i.e., a phenomenon that is also common in NN models.

The parameters' influence on the model's output can also be derived from the Jacobian matrix with respect to model parameters (see Supplementary Fig.~8 and Fig.~9). Here, a parameter’s influence on the model’s output refers to the sensitivity of the output with respect to changes in that parameter, as quantified by the Jacobian matrix. It is observed that the rotation vectors between the forearm and upper arm, the upper arm and shoulder, the shoulder and COM segments (i.e., $R^1_0$, $R^2_1$, and $R^3_2$), the relative position vectors between forearm and upper arm, the upper arm and shoulder (i.e., $P^1_0$ and $P^2_1$), have the highest influence on the model's output across all training epochs. The angular velocity coupling vector from the forearm to the upper arm (i.e., $A^1_0$) has slightly less influence. However, it is still higher than that from the upper arm to the shoulder (i.e., $A^2_1$) and the relative position vector from the shoulder to COM (i.e., $P^3_2$). Taking the rank deficiency into account, the relatively low sensitivity associated with the angular velocity coupling vector from the upper arm to the shoulder (i.e., $A^2_1$) and the relative position vector from the shoulder to COM (i.e., $P^3_2$) suggests that these parameters may be weakly identifiable or partially redundant. However, the observed rank deficiency may also arise from collinearity among model parameters as mentioned before.

\subsubsection{Physical interpretation of parameters}
\textbf{Independently-trained KM:} The activity-specific quaternions, representing effective segment-to-segment orientations (i.e., $R^1_0, R^2_1, R^3_2$), exhibit clear task-dependent differences. During natural walking, the mean rotation is modest ($\approx$ 41°), whereas balance- and turn-intensive tasks such as slalom walking, narrow beam walking, uneven terrain walking, and walk-and-turn are associated with substantially larger angles ($\approx$ 70–170°). 

The associated position vectors (i.e., $P^1_0, P^2_1, P^3_2$) remain within a consistent range of 0.1–0.3 m across tasks, reflecting plausible local segment offsets. The angular velocity coupling vectors (i.e., $A^1_0, A^2_1$) vary systematically: they are largest in beam and slalom walking, moderate in turning and uneven terrain, and minimal in sit-to-stand transitions. Together, these parameters reveal a coherent reweighting of orientation and coupling across tasks, consistent with increased proximal involvement and medio-lateral control in balance-challenging conditions.\\
\textbf{KM in sim1-/sim2-:} The simultaneously learned KM parameters exhibit a consistent pattern across both the FCNN- and LSTM-based models, with their behavior primarily depending on the degree of physical supervision imposed during training. In sim1-, where the KM is directly supervised by the ground-truth output in addition to being coupled to the NN, the physical parameters remain active across all activities and retain some task-dependent variation. For both FCNN and LSTM, the effective rotation parameters remain relatively large, with mean rotation angles generally exceeding 120° across activities. The learned position vectors also retain appreciable magnitudes, although they are generally larger than the approximately 0.1–0.3~m observed for the independently trained KM, reaching approximately 0.5~m on average and, for some parameter–activity combinations, approximately 0.6–0.9~m. Compared with the independently trained KM, the rotation parameters also exhibit less pronounced task-specific differentiation. These observations suggest that simultaneous optimization with either NN architecture introduces additional flexibility into the physical parameterization, even when the KM remains directly supervised.

A more pronounced change is observed for sim2-, where the KM is constrained only through its agreement with the NN. For both FCNN and LSTM, several position and angular velocity coupling parameters are substantially reduced, with near-zero values occurring for some activities, particularly narrow-beam walking and sit-to-stand transitions. In contrast, the effective rotation parameters generally remain comparatively large. This selective reduction indicates that the KM parameters do not simply decrease uniformly under weaker physical supervision. Rather, some components become weakly represented while others remain active. The consistency of this behavior across FCNN and LSTM suggests that it primarily arises from the coupling strategy rather than the specific data-driven architecture.

Overall, the comparison among the independently trained KM and the sim1- and sim2- models indicates a trade-off between physical constraint and parameter flexibility. The independently trained KM exhibits the clearest physically interpretable parameterization, including segment offsets within approximately 0.1–0.3~m and pronounced activity-dependent differences in effective rotations. Direct supervision in sim1- partially preserves the contribution of the physical parameters, while allowing greater deviation from the independently trained KM. Removing this supervision in sim2- provides substantially greater flexibility, allowing some position and angular-velocity coupling parameters to approach negligible values. Thus, progressively relaxing the supervision of the physical component increases its freedom to adapt to the data-driven model, but at the expense of a consistently physically interpretable parameterization.

\subsection{Simultaneous learning loss coupling analysis}
\subsubsection{Simultaneous learning approach 1 (sim1-)}
The simultaneous learning framework consists of three loss terms: the mechanistic loss $L_h$ of the KM model, the data loss $L_f$ of the NN model, and their coupled loss $L_c$, which enforces agreement between the outputs of the two models.

\begin{figure}[H]
    \centering
    \includegraphics[width=\linewidth]{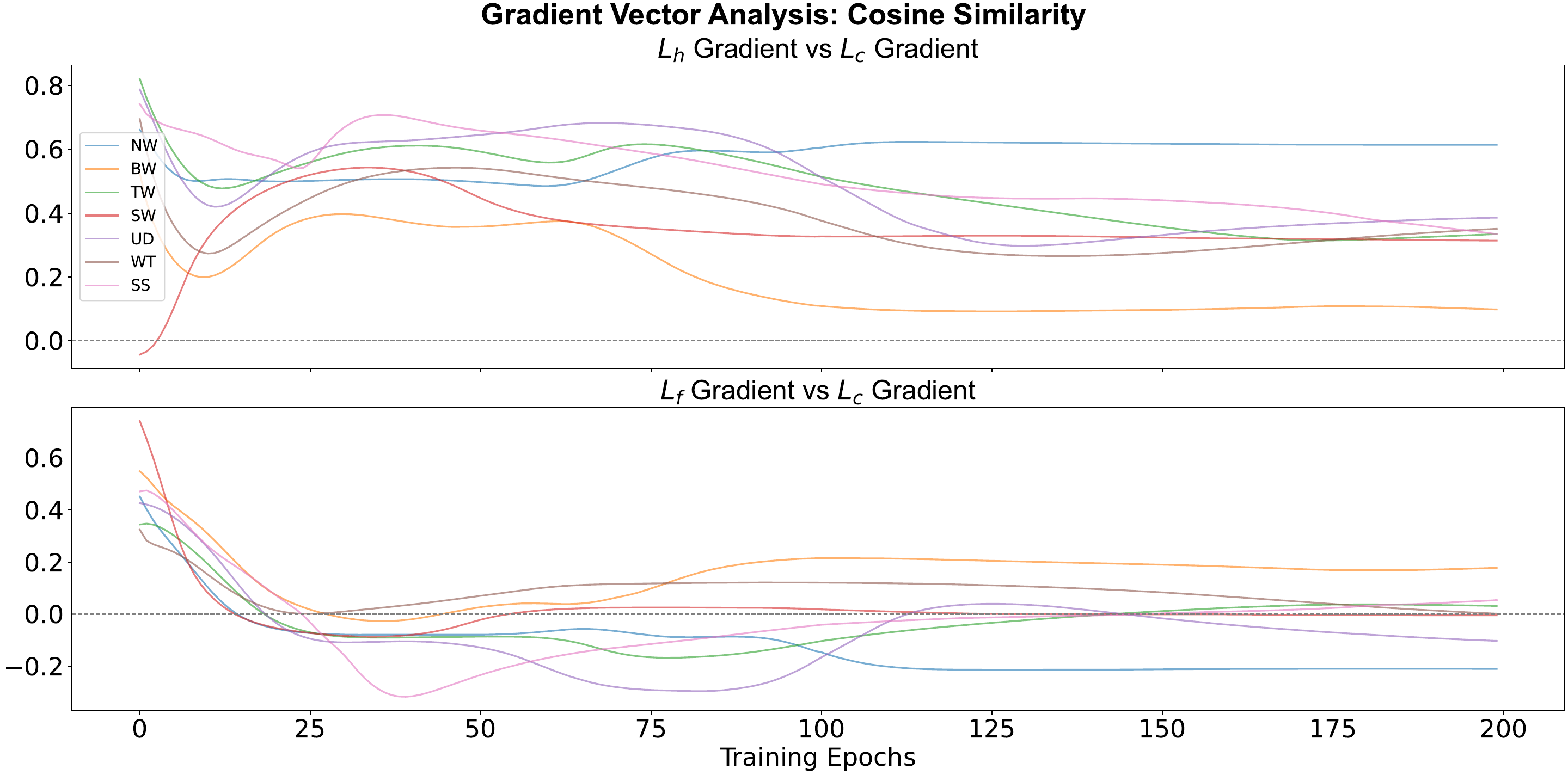}
    \begin{minipage}{0.8\linewidth}
            \footnotesize
            NW: natural walking; BW: beam walking; TW: terrain walking; SW: slalom walking; UD: upstairs and downstairs walking; WT: walk and turn; SS: sitting and standing up.
        \end{minipage}
        
    \caption{Simultaneous learning approach 1 (sim1-): Evolution of the cosine similarity between the gradients of the mechanistic loss $L_h$ and coupled loss $L_c$, and between the gradients of the neural network data loss $L_f$ and coupled loss $L_c$, across training epochs for different activities. Positive cosine similarity indicates cooperative optimization between loss terms, whereas negative values indicate gradient conflict.}
    \label{fig:cosine_similarity_PMB}
\end{figure}

The cosine similarity between two gradient vectors quantifies the optimization relationship between different loss terms during training. A positive cosine similarity indicates that the two objectives produce parameter updates in similar directions (i.e., cooperative optimization), whereas a negative value indicates gradient conflict, meaning that minimizing one loss opposes the other. Values near zero suggest a weak interaction between the objectives. The detailed derivation for the cosine similarity score is presented in ``Cosine similarity analysis for simultaneous learning" in Methods.

The evolution of the cosine similarity between the gradients of $L_h$ and $L_c$, and between the gradients of $L_f$ and $L_c$, over the training epochs is summarised in Figure~\ref{fig:cosine_similarity_PMB}. The cosine similarity between $\nabla L_h$ and $\nabla L_c$ remains predominantly positive across training, indicating that the coupled loss is generally compatible with the mechanistic objective and reinforces physiologically consistent optimization. In contrast, the cosine similarity between $\nabla L_f$ and $\nabla L_c$ exhibits larger fluctuations and occasional negative values, suggesting that the NN data loss sometimes conflicts with the coupling constraint. However, these fluctuations gradually diminish as training progresses, suggesting that the optimization relationship between the two objectives becomes more stable. For most activities, the cosine similarity subsequently approaches values near zero, indicating increasingly weak interaction between the two gradient directions, while a few activities retain moderately positive or negative alignment. Overall, these results suggest that the initially dynamic interaction between the data-fitting and coupling objectives gradually stabilizes, with cooperative optimization becoming weaker for most activities.

\subsubsection{Simultaneous learning approach 2 (sim2-)}
\begin{figure}[H]
    \centering
    \includegraphics[width=\linewidth]{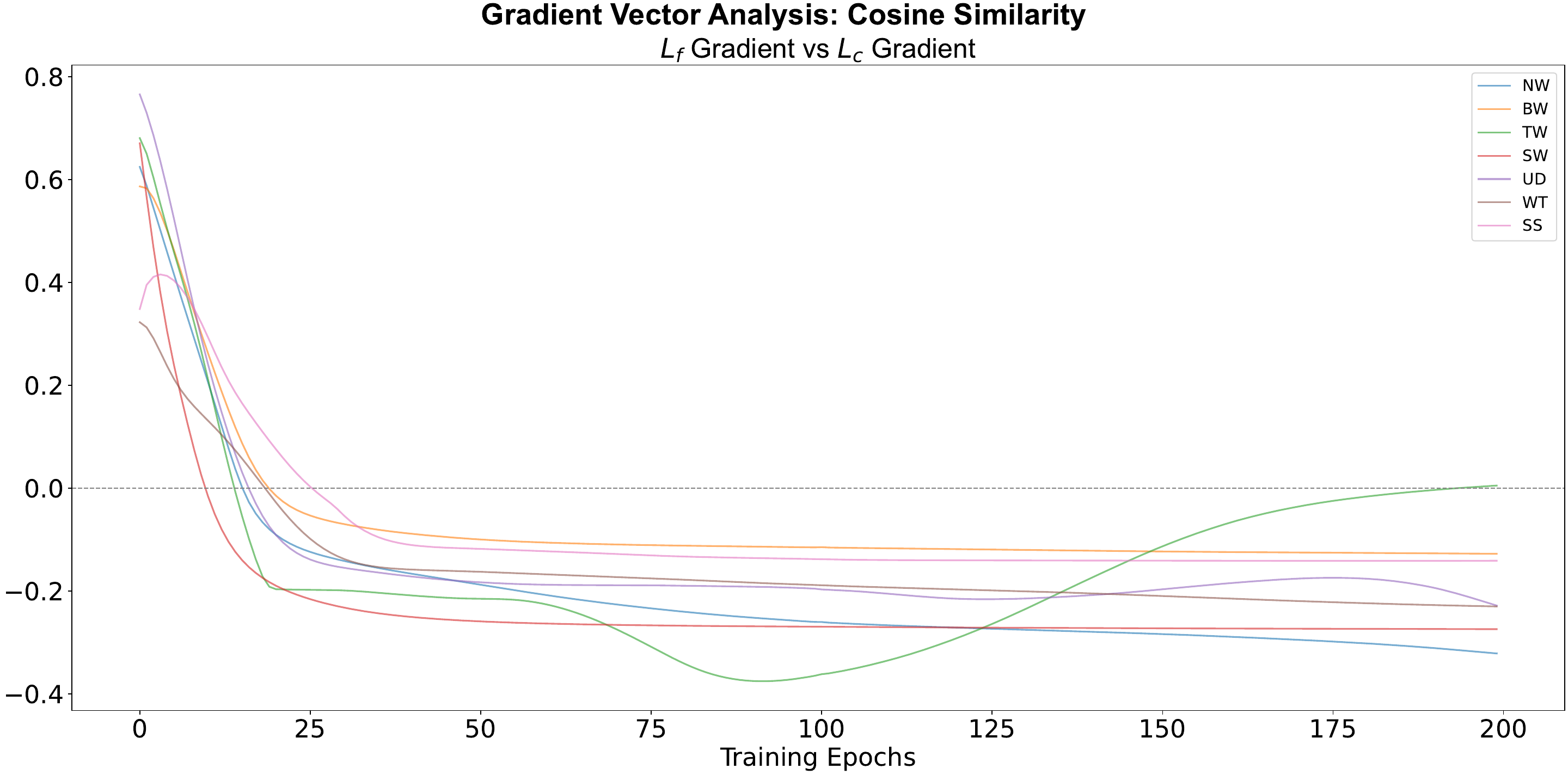}
    \begin{minipage}{0.8\linewidth}
            \footnotesize
            NW: natural walking; BW: beam walking; TW: terrain walking; SW: slalom walking; UD: upstairs and downstairs walking; WT: walk and turn; SS: sitting and standing up.
        \end{minipage}
        
    \caption{Simultaneous learning approach 2 (sim2-): Evolution of the cosine similarity between the gradients of the neural network data loss $L_f$ and coupled loss $L_c$, across training epochs for different activities. Positive cosine similarity indicates cooperative optimization between loss terms, whereas negative values indicate gradient conflict.}
    \label{fig:cosine_similarity_PMB2}
\end{figure}

As shown in Fig.~\ref{fig:cosine_similarity_PMB2}, the cosine similarity between the gradients of the data loss ($L_f$) and the coupled loss ($L_c$) exhibits a consistent trend across all activities. At the beginning of training, the cosine similarity is strongly positive (approximately 0.3–0.8), indicating that both objectives initially promote cooperative optimization directions. As training progresses, the similarity rapidly decreases and becomes slightly negative after approximately 20–30 epochs, stabilizing between -0.1 and -0.3 for most activities. Unlike the supervised simultaneous-learning approach (sim1-), the sim2- framework does not explicitly optimize the KM model using ground-truth COM acceleration. Instead, the physics branch is learned solely through its interaction with the NN via the coupled loss. Consequently, the gradual transition from positive to mildly negative cosine similarity suggests that the physics representation evolves together with the NN, initially learning features that support the data-fitting objective before gradually introducing biomechanical constraints that refine the learned representation. Importantly, the moderate magnitude of the negative cosine similarity indicates that the coupled loss functions as a soft regularizer rather than imposing conflicting optimization directions. An exception is the TW activity, where the cosine similarity gradually recovers toward zero during later training stages, suggesting that the learned physics representation and the data-driven objective become increasingly compatible as both branches co-adapt to capture the more complex dynamics of uneven-terrain locomotion.

\section{Methods}\label{sec11}
\subsection{Problem formualtion}
Estimating whole-body COM acceleration from wrist-worn inertial signals constitutes an ill-posed inverse problem. Formally, given wrist linear acceleration $a_0(t)$ and angular velocity $\omega_0(t)$, the goal is to recover the underlying COM acceleration $y_{com}(t)$ through an unknown nonlinear mapping, i.e.,

\begin{equation}
    y_{com}(t) = F(a_0(t), \omega_0(t))
\end{equation}

Because this mapping is inherently underdetermined, i.e., multiple whole-body configurations can produce similar wrist trajectories, additional physical priors are required to regularize the solution. We therefore formulate a hybrid physics–AI framework to address this inverse problem. A reduced-order KM provides interpretable biomechanical constraints on feasible wrist–COM mappings, while a neural learning operator $N_{\theta_R}$ complements it by capturing residual or unmodeled nonlinear dynamics. Depending on the integration strategy, the physics model (KM) acts either as a structural prior (ser-) or as a soft constraint in the optimization process (sim1- and sim2-). The resulting operator $F_{\theta_R, \theta_h}$ is trained to minimize a combination of physics-consistency and data-fidelity, effectively balancing interpretability and flexibility. The specific implementations of these three hybrid AI schemes are described in Section~\ref{sec:HKM-NN}.

\subsection{Reduced-order (4-segment KM) model}
\subsubsection{Model structure and formulation}
Our 4-segment (KM) model consists of the forearm segment, the upper arm segment, the shoulder segment, and the COM segment. According to the existing literature, the pelvis~\cite{b14, b16} or L5~\cite{b11, b12, b13} are two segments whose biomechanical centrality aligns with the physical location of COM and hence have been used as a proxy for COM measurements. Inspired by this finding, our KM model assumes the COM as a hypothetical segment, i.e., the last segment the linear accelerations are propagated to. Our KM model is inspired by kinematic models designed for robotic arm control~\cite{b17}, which is based on three of the Newton-Euler equations of motion. The three equations describe the propagation of the linear acceleration ($a$), angular velocity ($\omega$), and angular acceleration ($\dot{\omega}$) between consecutive body segments:

\begin{equation} \label{eq:acc}
\begin{split}
 a_{i+1} & = R^{i+1}_i[\dot{\omega}_i \times P^{i+1}_i + \omega_i \times (\omega_i \times P^{i+1}_i) + a_i]\\
 \omega_{i+1} & = R^{i+1}_i\omega_i + (\dot{\theta}_{i+1}Z_{i+1})\\
 \dot{\omega}_{i+1} & = R^{i+1}_i\dot{\omega}_i + R^{i+1}_i\omega_i \times (\dot{\theta}_{i+1}Z_{i+1}) + \Ddot{\theta}_{i+1}Z_{i+1}\\
\end{split}
\quad \text{for } i = 0, 1, 2, 3
\end{equation}

where $i$ denotes the numerical index of the segments, corresponding to forearm (i = 0), upper arm (i = 1), shoulder (i = 2), and COM (i = 3). $a \in \mathbb{R}^{3 \times n}, \omega \in \mathbb{R}^{3 \times n}, \dot{\omega} \in \mathbb{R}^{3 \times n}, n \in \mathbb{N}$ denote the linear acceleration, angular velocity, and angular acceleration, respectively. $R^{i+1}_i$ denotes the rotation matrix between segment $i$ and segment $i+1$. To represent the rotation matrix $R^{i+1}_i$, we use the quaternion. Thus, $R^{i+1}_i \in \mathbb{R}^4$ consists of four parameters $w$, $x$, $y$, and $z$. The rotation matrix $R^{i+1}_i$ transforms the coordinate system from frame $i$ to frame $i+1$, where a frame refers to a coordinate system attached to the segment. Hereby the same numerical index is used for the frame and the segment it is attached to. $P^{i+1}_i \in \mathbb{R}^3$ represents the position vector of segment $i+1$ relative to the segment $i$. Single and double dot notation represents first and second derivatives with respect to time. $Z_{i+1} \in \mathbb{R}^3$ is a unit vector that denotes the axis around which the joint $i+1$ rotates. $\dot{\theta}_{i+1}Z_{i+1} \in \mathbb{R}^3$ thus represents the angular velocity of joint $i+1$. $\Ddot{\theta}_{i+1}Z_{i+1} \in \mathbb{R}^3$ represents the angular acceleration of joint $i+1$. The symbol $\times$ denotes the cross-product operation.

Since only one IMU is attached to the wrist and there is no direct measurements of the joint angular velocities ($\dot{\theta}_{i+1}Z_{i+1}$) and angular accelerations ($\Ddot{\theta}_{i+1}Z_{i+1}$), our KM model approximates the propagation of the angular velocity from segment $i$ to segment $i+1$ as a linear transformation. Such an approximation is based on the following assumptions: (1) the forearm and upper arm segments are joined by a hinge joint (i.e., the elbow), which has 1 degree of freedom. Considering during gait and sitting-and-standing-up activities, there is limited rotation movement, our KM model assumes a rigid connection between the upper arm and the forearm, i.e., the elbow joint is locked; (2) the entire shoulder girdle has a total of 6 degrees of freedom. During gait and sitting-and-standing-up activities, the primary rotation that occurs is flexion and extension, representing 1 degree of freedom. In order to uphold the collinearity assumption between the upper arm and the shoulder, our KM model again assumes a rigid connection between the upper arm and the shoulder. Such reductive assumptions pose limitations on the accuracy of our KM model, but are necessary due to missing measurements on the joints. The linear transformation equation is presented below:

\begin{equation} \label{eq:angularvel}
\begin{split}
 \omega_{i+1} & = A^{i+1}_i(R^{i+1}_i\omega_i)\\
\end{split}
\end{equation}

where $A^{i+1}_i \in \mathbb{R}^3$ is a time-independent linear transformation vector. Based on Equations~\ref{eq:acc} and Equation~\ref{eq:angularvel}, the propagation of the angular acceleration from segment $i$ to segment $i+1$ is calculated as follows:

\begin{equation} \label{eq:angularacc}
    \begin{split}
        \dot{\theta}_{i+1}Z_{i+1} & = (A^{i+1}_i - 1)R^{i+1}_i\omega_i\\
 \Ddot{\theta}_{i+1}Z_{i+1} & = \frac{d(\dot{\theta}_{i+1}Z_{i+1})}{dt} = \frac{d((A^{i+1}_i - 1)R^{i+1}_i\omega_i)}{dt}\\
 \dot{\omega}_{i+1} & = R^{i+1}_i\dot{\omega}_i + [R^{i+1}_i\omega_i]_\times((A^{i+1}_i - 1)R^{i+1}_i\omega_i) \\
 & + \frac{d((A^{i+1}_i - 1)R^{i+1}_i\omega_i)}{dt}\\
    \end{split}
\end{equation}

where the derivative component is approximated with finite differences:

\begin{equation}
    \frac{d((A^{i+1}_i - 1)R^{i+1}_i\omega_i)}{dt} \approx (A^{i+1}_i - 1)R^{i+1}\frac{\omega_i - \omega_{i-1}}{\Delta t}
\end{equation}

where $\Delta t$ is equal to the sampling time interval.

Based on Equations~\ref{eq:acc}, Equation~\ref{eq:angularvel}, and Equations~\ref{eq:angularacc}, traversing from the forearm, upper arm, shoulder, up to the COM segment, our KM model comprises a total of 27 parameters to identify. These parameters include the rotation quaternion vector, the position vector, and the angular velocity coupling vector between the wrist and upper arm, denoted by $R^1_0 = (w^1_0, x^1_0, y^1_0, z^1_0)$, $P_0^1 = (px_0^1, py_0^1, pz_0^1)$, and $A_0^1 = (ax_0^1, ay_0^1, az_0^1)$, and those between the upper arm and shoulder, denoted by $R^2_1 = (w^2_1, x^2_1, y^2_1, z^2_1)$, $P^2_1 = (px_1^2, py_1^2, pz_1^2)$, and $A^2_1 = (ax_1^2, ay_1^2, az_1^2)$, and finally the rotation quaternion vector and the position vector between the shoulder and COM segment, denoted by $R^3_2 = (w^3_2, x^3_2, y^3_2, z^3_2)$ and $P^3_2 = (px_2^3, py_2^3, pz_2^3)$. These 27 parameters are indexed from 0 to 26 in this work.

It is worth noting that the acceleration and gyration from the forearm segment are sensor acceleration and segment gyration, measured by the Xsens MVN Link motion capture system, respectively. The reference COM acceleration is derived from the Xsens MVN Link motion capture system.

\subsubsection{KM model identification}
The KM model's parameters can be identified using the gradient descent method, implemented in the deep learning framework (PyTorch), for 125 epochs until convergence. 

The total mechanistic loss of the KM model is composed of two loss components, the acceleration loss $L_a$ and the quaternion loss $L_q$, defined as:

\begin{equation}\label{eq:L_h}
    L_{h} := L_a + L_q
\end{equation}

$L_a$ corresponds to the loss propagated from the wrist to the COM segment. It is defined as follows:

\begin{equation} \label{eq4}
 L_{a} := MSE(y_{com}(t), h(x_0(t); \theta_h))
\end{equation}

where $MSE()$ denotes the mean squared error function, $y_{com}(t) \in \mathbb{R}^{3 \times n}, n \in \mathbb{N}$ denotes the true values of the triaxial acceleration of the COM, and $h(x_0(t); \theta_{h}):\mathbb{R}^{9 \times n} \to \mathbb{R}^{3 \times n}, n \in \mathbb{N}$ denotes the output of the KM model, parametrized by $\theta_{h}$. The KM model takes the IMU data $x_0(t) \in \mathbb{R}^{9 \times n}, n \in \mathbb{N}$ from the wrist as input.

$L_q$ is applied to the model parameters comprising the three rotation matrices, $R^1_0$, $R^2_1$, and $R^3_2$. Each rotation matrix is represented by a quaternion (i.e., four parameters). $L_q$ imposes a constraint that the norm of the quaternion vector should be equal to 1. It is defined as follows:

\begin{equation} \label{eq:quaternion_loss}
\begin{split}
 L_{q} & := MSE(1, ||\theta_{h, R^1_0}||) + MSE(1, ||\theta_{h, R^2_1}||) + MSE(1, ||\theta_{h, R^3_2}||)\\
\end{split}
\end{equation}

where $||\theta_{h, R^1_0}||$ denotes the norm of the quaternion vector representing the rotation matrix between the wrist and the upper arm segments, $||\theta_{h, R^2_1}||$ between the upper arm and the shoulder segments, and $||\theta_{h, R^3_2}||$ between the shoulder and the COM segments.

\subsubsection{KM model parameter analysis}
The sensitivity and identifiability analysis of the KM model parameters are evaluated using the parameter-space Jacobian matrix. The rank of the parameter-space Jacobian matrix is used to assess the model parameters' identifiability. The sensitivity of the model's output to its parameters is evaluated by computing the gradient norm of each parameter, obtained from the parameter-space Jacobian matrix, as follows:

\begin{equation}
    I_{\theta_h(i)} = \left\| \frac{\partial f}{\partial \theta_h(i)} \right\|_2, \quad i = 1, 2, \dots, 27
\end{equation}

\subsection{Neural learning model}
Two NN model architectures are implemented: a long short-term memory model (LSTM) and a fully connected neural network (FCNN). For both NN model architectures, the Adam optimizer~\cite{b18} is used with a learning rate of 0.00005 and a weight decay of 0.01. The batch size and sequence length for our time-series IMU data during training are selected to be 1 and 256, respectively. The time stride between batches is selected to be half the sequence length, 128. In particular, for the LSTM layer, a stateful LSTM is implemented. This means that instead of being updated after every batch, the hidden and cell states of the LSTM layer are only updated after every 2200 data points, which corresponds to the duration of each measurement. 

\subsection{Human kinematic model-based neural networks}\label{sec:HKM-NN}
We propose three ensemble learning approaches to combine our KM model with neural learning to estimate COM accelerations: serial learning and two variants of simultaneous learning. 

\subsubsection{Serial learning}
\begin{figure}[!ht]
\centerline{\includegraphics[width=0.8\textwidth]{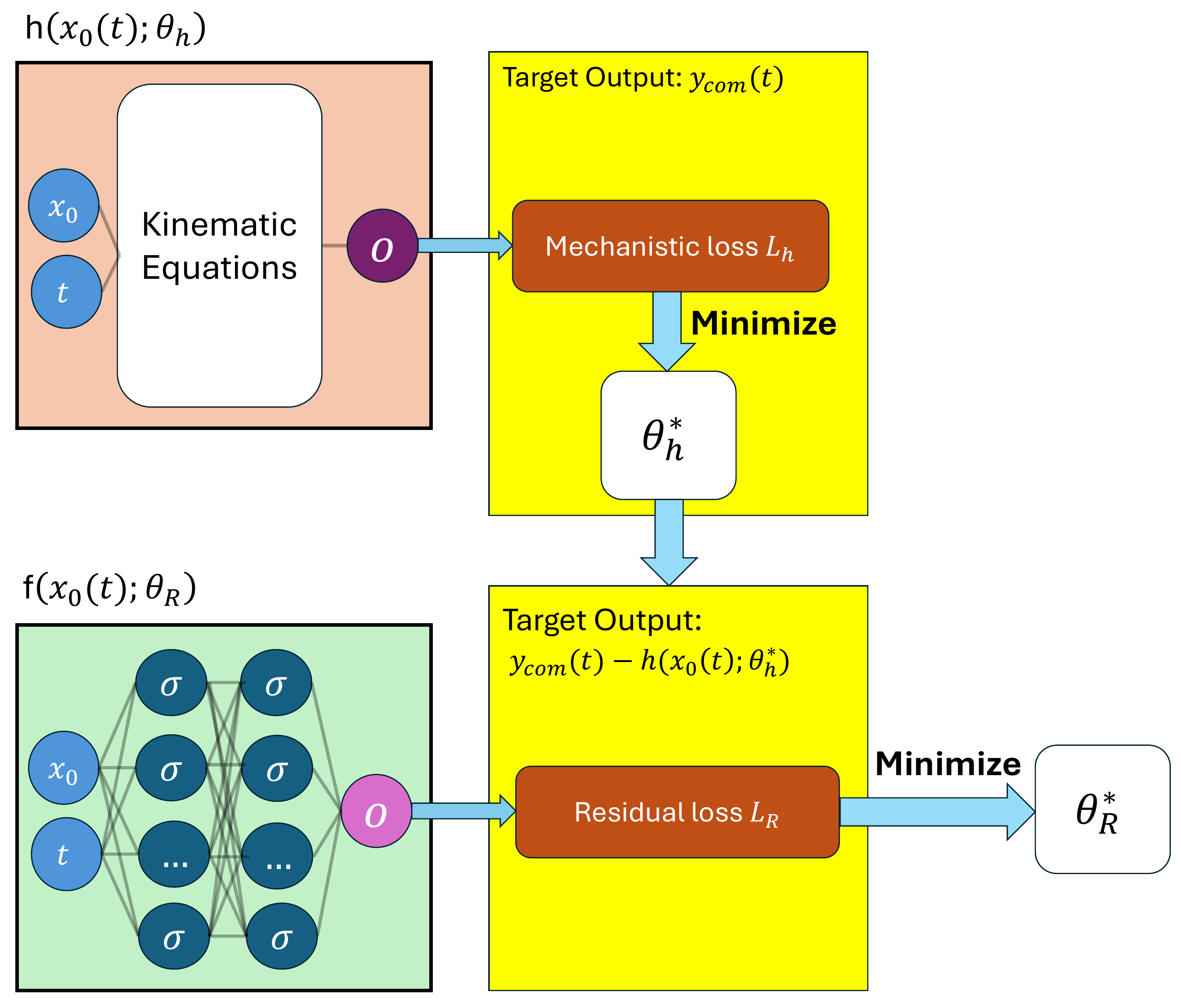}}
\caption{Diagram of serial learning. $h(x_0(t); \theta^*_h)$ represents the output of the trained KM model, where $\theta^*_h$ denotes its optimized parameters. $\theta^*_R$ denotes the optimized model parameters of the neural networks. }
\label{fig:seriallearning}
\end{figure}

The KM model is first trained to infer the unknown parameters $\theta_h$ using the gradient descent method, implemented in the deep learning framework (PyTorch), for 125 epochs until convergence. The output of the trained KM model is denoted as $h(x_0(t); \theta^*_h)$, where $\theta^*_h$ denotes its optimized parameters after convergence. Subsequently, an NN model receiving the same input data $x_0(t)$ is employed to learn the residuals between $h(x_0(t); \theta^*_h)$ and the target reference values (i.e., the COM accelerations). The optimized NN model parameters are denoted as $\theta^*_R$. Consequently, the framework retains the interpretability of the mechanistic model while using the NN only to compensate for unmodeled dynamics and structural mismatches. The NN therefore acts as a residual correction module rather than replacing the underlying physical interpretation of the KM model. The workflow is shown in Figure~\ref{fig:seriallearning}.

The residual loss of the NN model is defined as follows:

\begin{equation} \label{eq6}
\begin{split}
 L_{R} & := MSE(y_{com}(t) - h(x_0(t); \theta^*_h), f(x_0(t); \theta_R))\\
\end{split}
\end{equation}

where $h(x_0(t); \theta^*_h)$ denotes the output of the KM model with the inferred parameter set $\theta^*_{h}$ and $f(*; \theta_R):\mathbb{R}^{9 \times n}  \to \mathbb{R}^{3 \times n}, n \in \mathbb{N}$ denotes an NN model parametrized by $\theta_R$.

For the NN part of serial learning, we implement two NN models: fully-connected neural networks (FCNN) with one hidden layer of size 16 (ser-FCNN) and an LSTM layer with a hidden size of 8 (ser-LSTM). These hyperparameters are determined empirically through experimentation. For the FCNN, it is observed that increasing the hidden layer size beyond 16 does not yield substantial performance gains, while reducing it below 16 (e.g., to 8) leads to a notable decline in performance. Therefore, a hidden layer size of 16 is chosen as a balance between computational efficiency and model effectiveness. A similar observation and rationale guided the selection of a hidden size of 8 for the LSTM layer.

\subsubsection{Simultaneous learning}
Instead of training the KM and NN models sequentially, as in serial learning, both models are trained simultaneously. Specifically, the output of the KM model is used to constrain the NN during optimization through a coupled loss formulation, thereby encouraging the NN predictions to remain consistent with the mechanistic model. Two distinct approaches within this framework are proposed, i.e., the sim1- and the sim2-, where the sim1- is a further development of the sim2 which is originally proposed by Zhang et al.~\cite{zhang2026physiological}, as detailed below.

\textbf{Approach 1 (sim1-)}
\begin{figure}[!ht]
\centerline{\includegraphics[width=\textwidth]{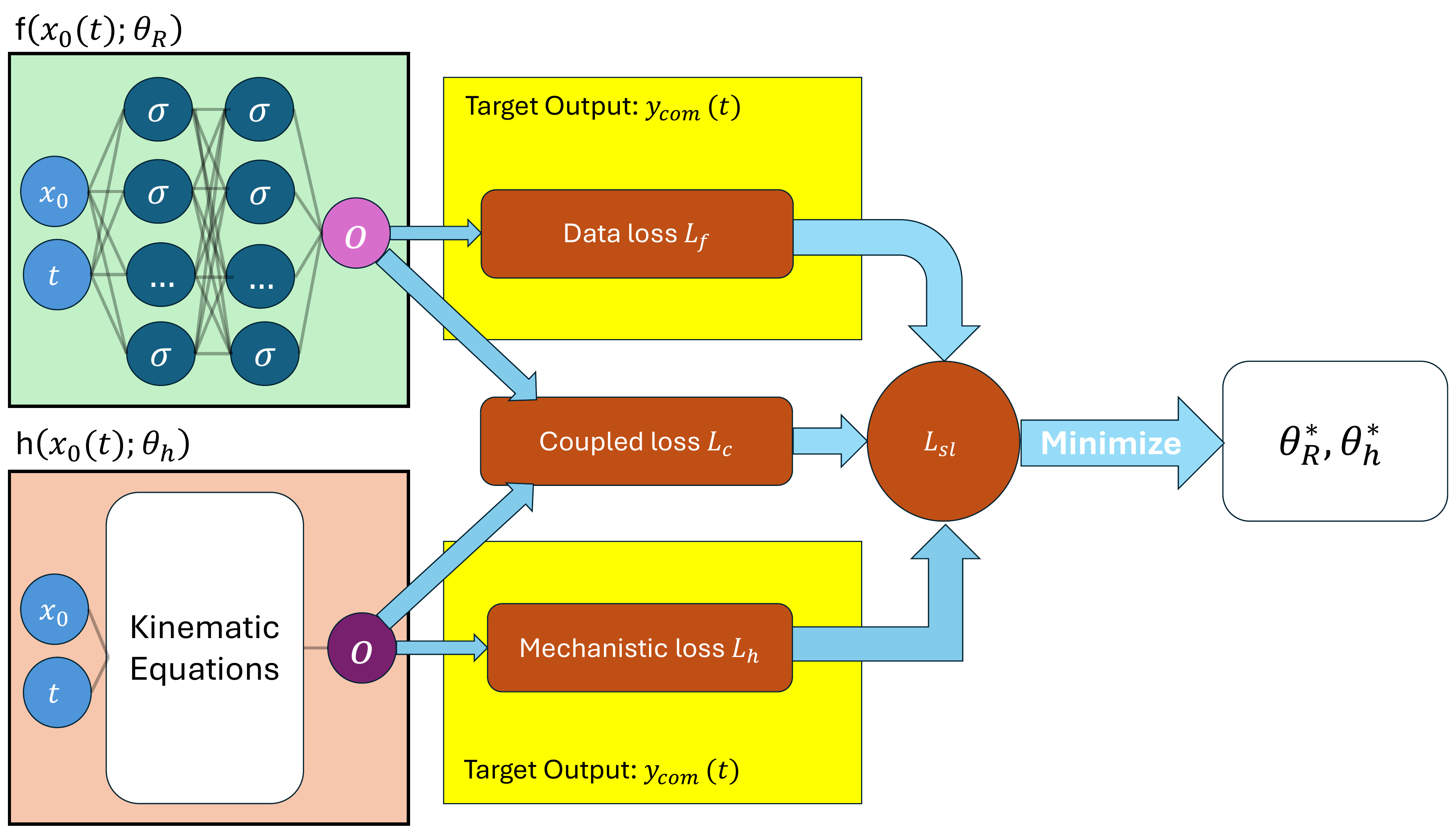}}
\caption{Diagram of simultaneous learning approach 1 (sim1-). $\theta^*_h$ denotes the optimized parameters for the KM model, and $\theta^*_R$ denotes the optimized parameters of the neural networks. Both $\theta_h$ and $\theta_R$ are optimized simultaneously through the three loss components $L_f$, $L_h$, and $L_c$, and $L_{sl}$ = $L_f$ + $L_h$ + 0.5$L_c$.}
\label{fig:physiolearning1}
\end{figure}

In this approach, the NN parameters $\theta_R$ are optimized using a data loss $L_f$, while the KM parameters $\theta_h$ are optimized using the total mechanistic loss $L_h$ (see Equation~\ref{eq:L_h}). The addition of $L_h$ is a further development based on the sim2- method (see the following section for details). Additionally, both $\theta_R$ and $\theta_h$ are jointly regularized through a coupled loss term $L_c$, which enforces consistency between the KM and NN predictions.

The data loss $L_f$ is defined as follows:

\begin{equation}
    L_f := MSE(y_{com}(t), f(x_0(t); \theta_R))
\end{equation}

The coupled loss term $L_c$ is defined as follows:

\begin{equation}
    L_c := MSE(f(x_0(t); \theta_R), h(x_0(t); \theta_h))
\end{equation}

For the NN model, the same FCNN and LSTM architectures and hyperparameter settings as those used in the serial learning framework are adopted.

The total simultaneous learning loss $L_{sl}$ is defined accordingly:

\begin{equation}
    L_{sl} = L_f + L_h + 0.5 \cdot L_c 
\end{equation}

where the weighting coefficient 0.5 is introduced to balance the influence of the coupled loss term relative to the data and mechanistic losses, preventing the consistency constraint from dominating the optimization process while still encouraging agreement between the KM and NN predictions.

\textbf{Approach 2 (sim2-)}
The principle of the sim2- method, as proposed by Zhang et al.~\cite{zhang2026physiological}, is similar to that of the sim1- method, with two distinctions. The first is the absence of the total mechanistic loss $L_h$, and the second is the equal weight between $L_f$ and $L_c$, i.e., $L_{sl} = L_f + L_c$. In this case, KM is not directly supervised by the ground truth but instead serves as a structural prior that guides the NN through the coupled loss $L_c$.

\begin{figure}[!ht]
\centerline{\includegraphics[width=\textwidth]{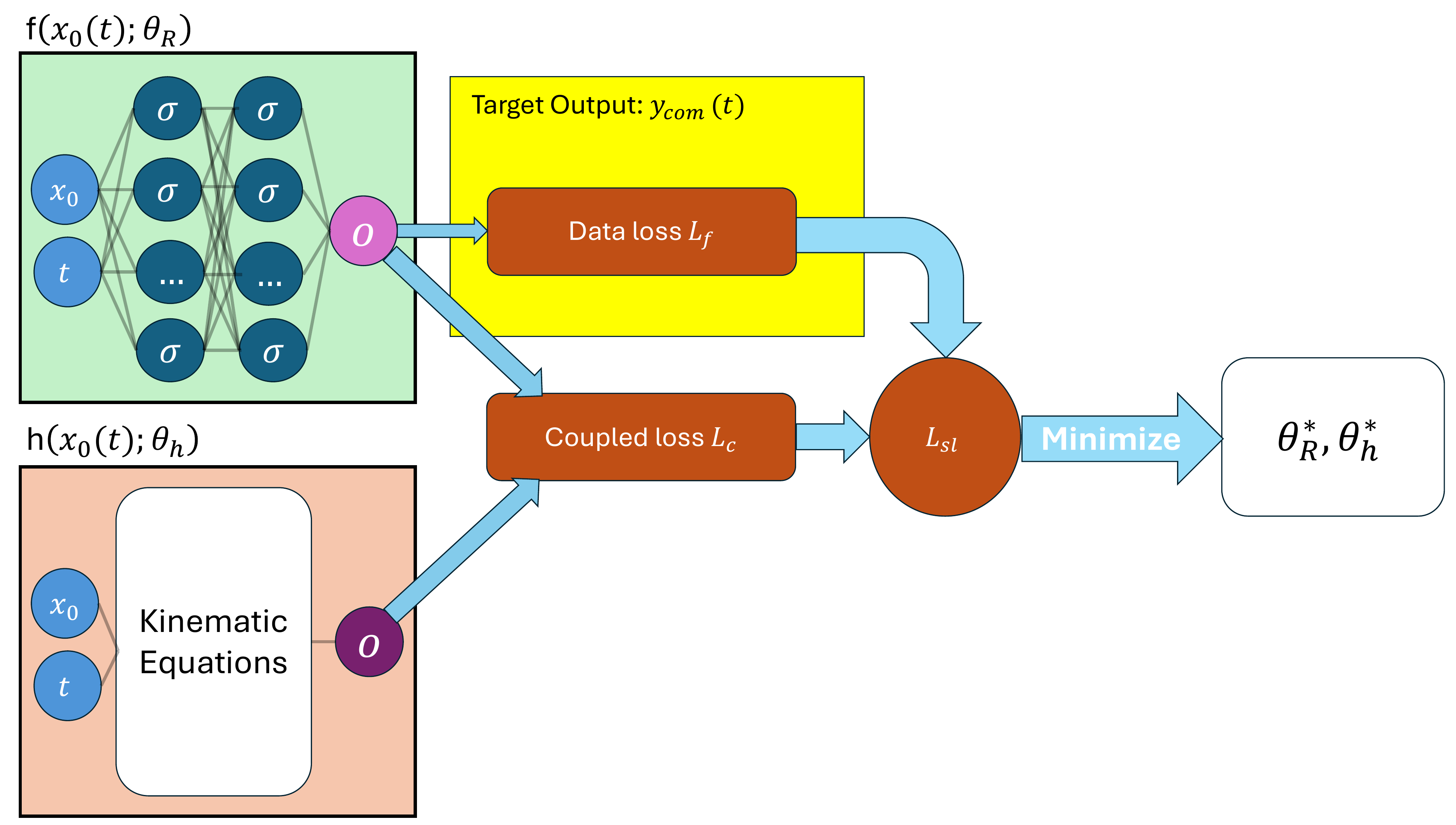}}
\caption{Diagram of simultaneous learning approach 2 (sim2-). $\theta^*_h$ denotes the optimized parameters for the KM model, and $\theta^*_R$ denotes the optimized parameters of the neural networks. Both $\theta_h$ and $\theta_R$ are optimized simultaneously through two loss components $L_f$ and $L_c$, and $L_{sl}$ = $L_f$ + $L_c$.}
\label{fig:physiolearning2}
\end{figure}

\subsubsection{Cosine similarity analysis for simultaneous learning}
To evaluate the co-optimization behavior of the mechanistic loss $L_h$ and the data loss $L_f$ relative to the coupled loss term $L_c$, the cosine similarity between their gradient vectors~\cite{schutze2008introduction} with respect to the model outputs is computed. 

The calculation of the cosine similarity score $CS_{fc}$ between the data loss gradient and the coupled loss gradient is presented below:

\begin{equation}
    CS_{fc} = 
\frac{
\nabla_{z_{fc}} L_f \cdot \nabla_{z_{fc}} L_{c}
}{
\left\| \nabla_{z_{fc}} L_f \right\|_2 
\cdot 
\left\| \nabla_{z_{fc}} L_{c} \right\|_2
}
\end{equation}

where $z_{fc}$ denotes the shared output representation with respect to which both losses $L_f$ and $L_c$ are differentiated.

The calculation of the cosine similarity score $CS_{hc}$ between the total mechanistic loss gradient and the coupled loss gradient is presented below:

\begin{equation}
    CS_{hc} = \frac{
\nabla_{z_{hc}} L_h \cdot \nabla_{z_{hc}} L_{c}
}{
\left\| \nabla_{z_{hc}} L_h \right\|_2 
\cdot 
\left\| \nabla_{z_{hc}} L_{c} \right\|_2
}
\end{equation}

where $z_{hc}$ denotes the shared output representation with respect to which both losses $L_h$ and $L_c$ are differentiated.

The cosine similarity score ranges from -1 to 1, where a positive value indicates positive alignment between two gradient vectors, i.e., cooperative learning, a negative value indicates negative alignment, i.e., competitive learning, and a value of $0$ indicates orthogonality, i.e., independent learning.

\subsection{Dataset and experimental protocol}
\begin{figure}[H]
    \centering
    \includegraphics[width=0.8\textwidth]{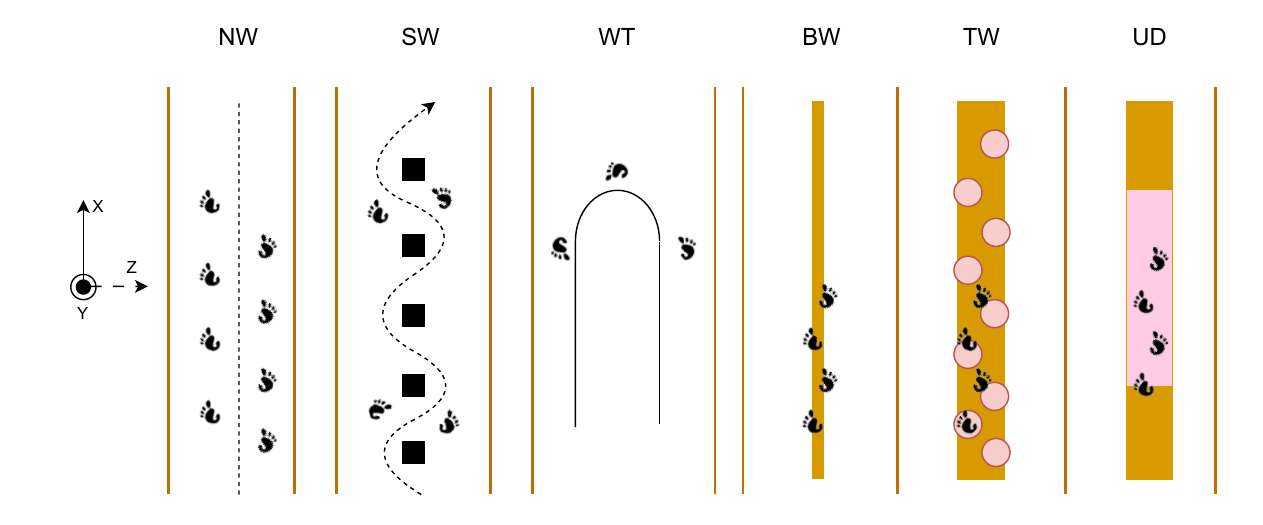}
    \caption{An illustration of all the gait activities, including natural walking (NW), slalom walking (SW), walk and turn (WT), narrow beam walking (BW), uneven terrain walking (TW), and upstairs and downstairs walking (UD). The pink circles of the TW indicate places where the surface protrudes, resulting in an uneven surface. The coordinate system during measurement is presented on the left, where the x-axis, y-axis, and z-axis denote the anterior–posterior, vertical, and medio-lateral directions, respectively. The pink rectangle of the UD indicates the raised platform after walking upstairs. The stairs are not explicitly shown here.}
    \label{fig:gaits}
\end{figure}

The research was conducted at Roessingh Research and Development, Enschede. The study (File no. 230257) was approved by the Ethics Committee Computer \& Information Science of the University of Twente. In total, IMU data from 10 healthy volunteers were successfully collected, including four females and six males with an age of 27 years old, a mean (standard deviation) weight of 72.4 (10.8) kg, and a mean (standard deviation) height of 1.73 (0.12) m. Informed consent was obtained from each participant before their respective data collection. The participants were asked to wear the Xsens MVN Link motion capture suit (17 IMUs) to perform a list of body movement activities. In this work, we used the IMU attached to the left forearm, i.e., to the wrist, as the source of our wrist-worn IMU measurements. As for the reference COM accelerations, we used the COM output calculated by the Xsens Link system. The list of body movement activities includes natural walking (NW), slalom walking (SW), sitting down and standing up (SS), walking and turning (WT), walking on a narrow beam (BW), walking on uneven terrain (TW), up- and down-stairs walking (UD). An illustration of all the gait activities is shown in Figure\,\ref{fig:gaits}.

\subsection{Evaluation methods}
\subsubsection{Evaluation metrics} To quantify the agreement between the reference COM linear acceleration measurements $(y_{com}, t)$ produced by the Xsens Link system and our predictive models $(\hat{y}_{com}, t)$, three metrics are used, including the normalized root-mean-squared error (NRMSE), coefficient of determination ($R^2$), and Pearson's correlation coefficient ($r$)~\cite{b19}.

The NRMSE is defined as follows:

\begin{equation} \label{eq8}
\begin{split}
 NRMSE & = \frac{\sqrt{\frac{1}{n} \sum_{t=1}^n (x_t - y_t)^2}}{\max(y) - \min(y)} \times 100\\
\end{split}
\end{equation}

where n denotes the number of data points, i.e., sampled time steps, $x_t$ denotes the predicted values, and $y_t$ denotes the true values. 

The coefficient of determination ($R^2$) is calculated as follows:

\begin{equation} \label{eq10}
\begin{split}
    R^2 & = 1 - \frac{\sum_{t=1}^n (y_t - x_t)^2}{\sum_{t=1}^n (y_t - \bar{y})^2}
\end{split}
\end{equation}

where $y_t$ denotes the true values, $x_t$ the predicted values, and $\bar{y}$ the mean of the true values.

The Pearson's correlation coefficient $r$ is calculated as follows:

\begin{equation} \label{eq9}
\begin{split}
    r & = \frac{\sum_{t=1}^n (x_t - \bar{x})(y_t - \bar{y})}{\sqrt{\sum_{t=1}^n (x_t - \bar{x})^2} \sqrt{\sum_{t=1}^n (y_t - \bar{y})^2}}
\end{split}
\end{equation}

where $\bar{x}$ and $\bar{y}$ denote the mean values of $x_t$ and $y_t$, respectively.

The NRMSE expresses the prediction error as a percentage of the signal range, with lower values indicating better agreement. An NRMSE of $0\%$ denotes a perfect fit, while higher values reflect greater deviations between the predicted and reference signals. Higher values of $R^2$ indicate that a larger proportion of the variance in the reference signal is explained by the model, with $R^2 = 1$ denoting a perfect fit and values below zero indicating performance worse than a mean predictor ($R^2 = 0$). The Pearson’s correlation coefficient $r$ reflects the strength and direction of the linear association between predicted and true values, ranging from $-1$ (perfect negative correlation) to $+1$ (perfect positive correlation), with $r = 0$ denoting no linear correlation.

\subsubsection{Baseline models for comparison} 
To assess whether our proposed HKM-NN models offer performance improvements, we use two baseline models for comparison: 1) our proposed KM model as used for the mechanistic component used in the HKM-NN; 2) pure data-driven NN models, including one LSTM model ($n_h = 8$) and one FCNN model ($n_h = 16$) as used for the pure NN component used in the HKM-NN. 

\subsubsection{Influence of artificial noise}
To model uncertainty and measurement corruption encountered in real-world wearable sensing, Gaussian noise $N(\mu, \sigma^2)$ is injected into the test input signals during evaluation. Gaussian noise represents continuous stochastic perturbations commonly arising from sensor electronics, thermal noise, quantization effects, soft-tissue motion artifacts, and imperfect sensor attachment. In addition to Gaussian noise, salt-and-pepper noise~\cite{b24} is introduced to simulate occasional gross signal corruption, such as abrupt spikes, dropouts, packet transmission errors, transient sensor displacement, or intermittent loss of skin contact that frequently occur during unconstrained daily-life monitoring. 

To evaluate each model’s sensitivity to varying levels of signal degradation, multiple signal-to-noise ratio (SNR) conditions are considered during testing, including SNR values of 2, 4, 6, 8, and 10. Lower SNR values correspond to more severe signal corruption and, therefore, more challenging sensing conditions.

Importantly, the noise is applied only during testing, while all models are trained using clean supervision. This setting intentionally introduces a distribution shift between training and deployment conditions, thereby evaluating the robustness and generalization capability of each model under realistic out-of-distribution wearable sensing scenarios. Such evaluation is particularly relevant for real-world digital health applications, where wearable measurements acquired during activities of daily living are often substantially noisier and less controlled than laboratory-collected training data.

\subsubsection{Statistical test}
The non-parametric repeated measures Friedman test is used to evaluate the statistical significance of the differences among different models and activities. The Wilcoxon signed-rank test is used to perform non-parametric post hoc tests with Bonferroni correction. These two tests do not require the normality of the data.

\section{Discussion}\label{sec12}
The obtained results demonstrate the feasibility of estimating COM acceleration from a single wrist IMU across multiple gait activities and the sit-to-stand transition task. Furthermore, the comparative analysis indicates that model performance depends strongly on both the hybrid-AI architecture and the type of noise perturbation present in the test data.

The results reveal complementary limitations of the purely mechanistic and purely data-driven models. The standalone KM provides the most transparent biomechanical parameterization, with activity-dependent orientations and plausible segment offsets, but its sub-full-rank Jacobian indicates that parts of the simplified kinematic chain are weakly identifiable or redundant. Its sensitivity to Gaussian perturbations further shows that physical structure alone does not guarantee robustness when measurement errors continuously affect the quantities propagated through the model. Conversely, the FCNN and LSTM can learn relationships beyond the simplified kinematic formulation and therefore achieve stronger predictive performance, but their robustness depends on their capacity to process temporal disturbances. In particular, the greater stability of the LSTM under sparse impulsive noise suggests that temporal context can reduce sensitivity to isolated corrupted samples. The comparison between these two model classes therefore establishes the central challenge addressed by hybridization, i.e., retaining useful physical structure without allowing the limitations of the mechanistic formulation to dominate prediction.

The serial framework represents the most explicit form of physical integration because the NN prediction depends directly on the intermediate output of the KM. This physical bottleneck provides a clear functional role for the KM and can be beneficial when its representation remains stable, as suggested by the comparatively strong behavior of ser-FCNN under salt-and-pepper noise. In this case, transforming the corrupted measurements through the KM may partly compensate for the FCNN’s lack of explicit temporal modeling. However, the same dependency becomes a limitation under Gaussian noise, where perturbations affecting the KM output are transmitted to the downstream NN and cannot necessarily be corrected. The serial approach therefore prioritizes a clearly defined physical information pathway, but its robustness is bounded by the reliability of the mechanistic representation under the encountered perturbation.

The sim1- framework achieves a different balance by directly supervising both the predictive and kinematic components while encouraging agreement between them. Its comparatively strong predictive performance across clean and Gaussian-noise conditions indicates that this strategy can preserve data-driven adaptability without making the NN deterministically dependent on the KM output. The parameter analysis nevertheless shows that the jointly learned KM does not retain the same degree of physical interpretability as the independently trained KM. For example, its position parameters become larger, and its rotations show less activity-specific differentiation. Thus, direct supervision keeps the physical branch active and task-responsive, but does not prevent its parameters from adapting to the joint optimization objective. The gradient analysis supports this interpretation at the optimization level. The mechanistic and coupling losses remain predominantly cooperative, whereas the interaction between the NN data loss and the coupling loss becomes more stable during training and approaches weak alignment for most activities, while retaining moderate cooperation or conflict for others. The sim1- should therefore be understood not as enforcing a fixed biomechanical rule, but as maintaining an explicitly supervised physical reference within a flexible joint solution.

The sim2- demonstrates the consequences of relaxing this physical supervision further. Its lower predictive accuracy, together with the reduction of several position and angular velocity coupling parameters, indicates that agreement between the two branches can be achieved without preserving the full biomechanical contribution of the KM. At the same time, its performance often changes relatively little as noise increases from its already lower noise-free baseline. This behavior should not be interpreted as evidence that the KM provides robust biomechanical guidance, because the standalone NN models achieve comparable or better accuracy without such a prior. Instead, the combined evidence raises the possibility that weakly supervised coupling favors a more constrained solution that is less responsive to both informative input variation and introduced perturbations. The transition of the data–coupling gradient relationship from initial cooperation to mild conflict is consistent with an optimization process in which agreement with the physical branch increasingly competes with predictive fitting. However, the present analyses cannot establish whether the limited noise-induced degradation arises from parameter rigidity, reduced input sensitivity, or other characteristics of joint optimization. This mechanism therefore requires further investigation.

Together, the experiments show that hybridization does not produce a single, universal trade-off between accuracy, robustness, and interpretability. Rather, each integration strategy determines how errors and constraints are transmitted through the system. Serial learning gives the physical model a clear and interpretable computational role, but can propagate its sensitivity to the prediction stage. The sim1- provides the most balanced behavior in the present study by combining relatively high predictive accuracy, resistance to Gaussian perturbations, and an active though adapted physical branch. The sim2- offers greater freedom for co-adaptation, but this freedom can weaken the physical meaning of the learned parameters and may produce low-sensitivity solutions without improving accuracy. More broadly, these findings indicate that the value of a biomechanical prior depends not simply on whether it is included, but on how its physical meaning is maintained during optimization and how strongly the final prediction depends on it.

From an application perspective, the appropriate hybrid architecture may therefore depend on the expected sensing environment and the intended role of the physical model. A serial structure may be suitable when the mechanistic transformation is sufficiently reliable and physical interpretability is important, whereas supervised simultaneous learning (sim1-) may be preferable when predictive performance and tolerance to distribution shift must be balanced. If physical interpretability is a central objective, the behavior of sim2- indicates that agreement between the KM and NN outputs alone is insufficient to preserve a physically interpretable mechanistic representation. Although direct KM supervision in sim1- helps maintain the contribution of the physical parameters, additional parameter constraints or identifiability-aware regularization may still be required to preserve their biomechanical meaning during joint optimization. Future work should consequently evaluate not only prediction accuracy under perturbation, but also changes in input sensitivity, parameter identifiability, and physical consistency across training and deployment conditions.

\section{Conclusions}
Our results demonstrate the feasibility of estimating COM acceleration from a single wrist-worn IMU across multiple gait activities and the sit-to-stand transition task. The proposed KM model provides an interpretable physics-based representation of movement dynamics, while the NN and most HKM-NN models further improve predictive accuracy by capturing nonlinear relationships beyond the simplified kinematic formulation. In particular, sim1- achieves higher predictive accuracy and preserves a more active physical parameterization than sim2-, suggesting that direct supervision of the KM branch provides a more effective balance between physical constraint and data-driven adaptability than coupling-only simultaneous learning.

The results further show that robustness depends on both the hybrid learning strategy and the characteristics of the noise. Under Gaussian noise, sim1- maintains comparatively high predictive accuracy, whereas sim2- exhibits lower accuracy but relatively limited additional degradation from its noise-free baseline. Under salt-and-pepper noise, the KM remains comparatively stable, while the differences between FCNN- and LSTM-based models indicate that temporal modeling contributes to resistance against sparse impulsive perturbations. The cosine similarity analysis further demonstrates that joint learning involves dynamically changing cooperative and conflicting interactions between the loss objectives, with these optimization dynamics depending on whether the KM branch is directly supervised or constrained solely through output coupling.

Overall, these findings demonstrate the potential of hybrid-AI approaches for wearable motion analysis, while showing that predictive accuracy, robustness, and physical interpretability are not automatically improved together. Their balance depends on how the mechanistic and data-driven components are integrated and supervised. Future work should extend the framework to more diverse daily-life activities, improve the identifiability and biomechanical plausibility of the mechanistic parameters, and investigate adaptive coupling strategies that regulate the interaction between physical constraints and data-driven optimization during training.

\backmatter

\bmhead{Acknowledgement}
We acknowledge Prof. Peter Veltink for his contributions in providing initial feedback on the kinemtic modeling part. Funded by the European Union (the HealthyW8 project). Views and opinions expressed are however those of the authors only and do not necessarily reflect those of the European Union or HaDEA. Neither the European Union nor the granting authority can be held responsible for them.

\bmhead{Supplementary information}
A supplementary file is provided for this article.

\end{document}